%% file: acl_latex.tex
\documentclass[11pt]{article}

\usepackage[final]{acl}
\usepackage{microtype}
\usepackage{graphicx}
\usepackage{subcaption}
\usepackage{booktabs} 
\input{math_commands.tex}

\input{variables}
 \usepackage{multirow} 
 \usepackage{algorithm}
\usepackage{algpseudocode}
\usepackage{amsmath}
\usepackage{amsfonts}
\usepackage{nicefrac}
\usepackage{url}
\usepackage{times}
\usepackage{latexsym}

\usepackage[T1]{fontenc}

\usepackage[utf8]{inputenc}

\usepackage{microtype}

\usepackage{inconsolata}

\usepackage{amsthm}

\usepackage{graphicx}
\theoremstyle{plain}

\theoremstyle{definition}

\theoremstyle{remark}

\usepackage{listings}
\usepackage{tcolorbox}
\tcbuselibrary{listings} 
\tcbuselibrary{skins} 
\tcbuselibrary{listingsutf8}
\usepackage{xcolor}
\newtcblisting{mycodebox}{
  colback=gray!10,    
  colframe=gray!40,   
  listing only,
  listing options={
    basicstyle=\ttfamily\footnotesize,
    language=python,
    showstringspaces=false,
    breaklines=true,
  },
  left=0pt,
  right=0pt,
  top=2pt,
  bottom=2pt,
  boxsep=4pt,
  arc=4pt,
  enhanced,
}

\usepackage{arydshln}  

\usepackage{times}
\usepackage{latexsym}
\usepackage{multicol}
\usepackage{tabularx}
\usepackage{booktabs}
\usepackage{makecell}  
\usepackage{color}
\usepackage{xcolor}
\usepackage{orcidlink}

\newtcolorbox{mypromptbox}{colback=blue!5!white, colframe=blue!35!white, fonttitle=\bfseries, coltitle=black, title=Prompt}
\newtcolorbox{myleanbox}{colback=orange!5!white, colframe=orange!25!white, fonttitle=\bfseries, coltitle=black, title=}
\newtcolorbox{docstringbox}{colback=gray!5!white, colframe=gray!25!white, fonttitle=\bfseries, coltitle=black, title=Generated Docstring}
\usepackage{listings}
\usepackage{amssymb}

\definecolor{keywordcolor}{rgb}{0.7, 0.1, 0.1}   
\definecolor{tacticcolor}{rgb}{0.0, 0.1, 0.6}    
\definecolor{commentcolor}{rgb}{0.4, 0.4, 0.4}   
\definecolor{symbolcolor}{rgb}{0.0, 0.1, 0.6}    
\definecolor{sortcolor}{rgb}{0.1, 0.5, 0.1}      
\definecolor{attributecolor}{rgb}{0.7, 0.1, 0.1} 

\usepackage[textsize=tiny]{todonotes}

\title{TopoAlign: A Framework for Aligning Code to Math via Topological Decomposition}

\author{
    Yupei Li\thanks{Work conducted during an internship at Huawei Noah's Ark Lab, London.}\thanks{Equal contribution.} \orcidlink{0009-0003-0832-1984} \\ Imperial College London, UK \\ \texttt{yl7622@ic.ac.uk} \\
    \AND
    Philipp Borchert\footnotemark[2] \orcidlink{0000-0002-5533-4281} \\ Huawei Noah's Ark Lab, London, UK \\ \texttt{philipp.borchert@h-partners.com} \\
    \And
    Gerasimos Lampouras \orcidlink{0000-0002-4102-7256} \\ Huawei Noah's Ark Lab, London, UK \\ \texttt{gerasimos.lampouras@huawei.com} \\
}

\begin{document}
\maketitle
\begin{abstract}
Large Language Models (LLMs) excel at both informal and formal (e.g. Lean 4) mathematical reasoning but still struggle with \emph{autoformalisation}, the task of transforming informal into formal mathematical statements. Yet, the performance of current Math LLMs is constrained by the scarcity of large-scale corpora, particularly those containing pairs of informal and formal statements. Interestingly, the formal languages used in autoformalisation share structural similarities with programming languages, and code data is available at scale. However, current models trained on code do not transfer effectively to formal math, due to structural and syntactic differences between them. To address this, we propose \emph{TopoAlign}, a framework that unlocks widely available code repositories as training resources for Math LLMs. TopoAlign decomposes code into docstrings, main functions, and dependency functions, and reassembles these components into analogues that structurally mirror formal statements. We train three state-of-the-art models, \dsmath, \qwen and \herald, and evaluate them on the \minif, \putnam, and \proofnet benchmarks. TopoAlign provides substantial gains for \dsmath, improving performance by 17.77\% on BEq@10 and 68.82\% on typecheck@10, and also measurably improves \herald by 0.12\% on BEq@10 and 1.09\% on typecheck@10 despite introducing no new mathematical knowledge.
\end{abstract}
\section{Introduction}

Neuro-symbolic approaches that pair LLMs with proof assistants, such as Isabelle \citep{nipkow2002isabelle} or Lean 4 \citep{moura2021lean}, enable advanced math reasoning by enforcing rule-based logical consistency \citep{welleck2023llmstep}. These assistants operate on Formal Languages (FL), such as Lean 4. However, proficiency in these formal languages requires specialized expertise; consequently most math problems are initially expressed in Natural Language (NL). While NL is ideal for human communication, its inherent flexibility and contextual dependence make it challenging to translate into a formal system. Bridging this gap requires \emph{autoformalisation}, the process of faithfully translating informal NL math problems into FL. This step is essential for interacting with automated verifiers for tasks such as proof generation \citep{wu2022autoformalization, ahn2024large}.

Despite recent advances, LLMs still struggle with autoformalization, in part due to the lack of large-scale, high-quality parallel datasets pairing natural language problem descriptions with corresponding formal statements or proofs \citep{wu2022autoformalization}. Synthetic datasets such as Herald Statements \citep{gao2024herald}, Numina v1.5 ATF \cite{guo2025autoformalizer}, and Goedel-Pset \cite{lin2025goedelproverfrontiermodelopensource} partially address this gap, but their scale and diversity remain limited and not annotated by human expert. Furthermore, these approaches rely on generating synthetic informal-formal pairings, which are inherently constrained by the quality and coverage of the underlying generative models, and do not scale naturally with the availability of existing large-scale resources such as code repositories.
As a result, current models either fail outright or require thousands of attempts and auxiliary retrieval systems to produce accurate formalisations of even simple math problems \citep{li2024autoformalize}.

\begin{figure*}[t]
  \centering
  \includegraphics[width=1.8\columnwidth]{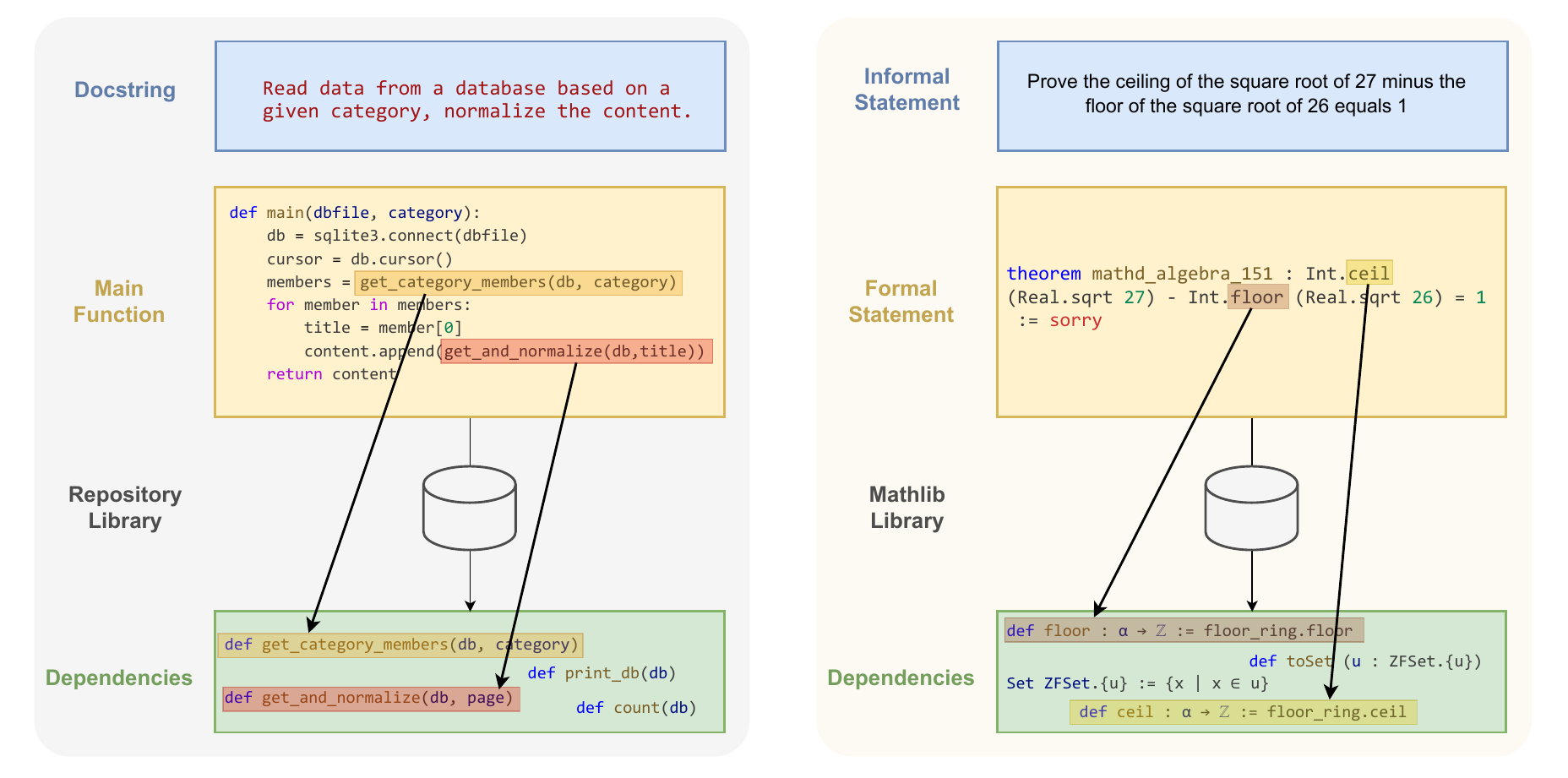}
  \caption{Structural similarity between code (left) and formal statements in Lean 4 (right). Code samples extracted from GitHub repositories are decomposed into: the docstring, which maps to informal statements in mathematical problems, the main function, which corresponds to the formal statements, and the dependency functions, which correspond to supporting lemmata and theorems, included in external libraries (e.g. Mathlib for Lean 4).}
  \label{fig:parallel}
  \vspace{-0.5cm}
\end{figure*}

To overcome this data bottleneck, we look to the vast resources available in software engineering. Building on the assumption that models can learn the \emph{structure} of a task from syntactically aligned data, even if the data is semantically distinct from the final domain. This insight unlocks widely available code repositories as large-scale data sources for training Math LLMs. Beyond providing scale, code data offers structural diversity, requiring complex problem-solving skills that we posit are transferable to formal mathematics. Leveraging these corpora requires aligning the compositional patterns of code with those of formal proofs.
We propose \emph{TopoAlign}, a framework that structurally aligns programming code with formal math. TopoAlign decomposes code into docstrings, main functions, and dependency functions, then reassembles these components into sequences that mirror the structure of Lean~4 formal statements (see Figure~\ref{fig:parallel}). This alignment teaches the model the compositional structure of formal math and enables the transfer of problem-solving capabilities learned from code without introducing new mathematical knowledge.
Using this aligned corpus, we introduce \emph{code autoformalisation} (CAF). This task emulates autoformalisation by aligning dependency functions, and main function bodies with informal descriptions (code docstrings), supporting lemmata, and formal statements in Lean. Unlike regular code generation, our setting provides a synthetic docstring that already includes the solution intent. We train \dsmath \citep{shao2024deepseekmath}, \herald \citep{gao2024herald}, and \qwen \citep{qwen3} with TopoAlign and the CAF objective. On the \minif, \putnam, and \proofnet benchmarks, the method yields consistent gains, achieving relative BEq improvements of 36.7\% for \dsmath and 6.2\% for \herald.

\textbf{Contributions:} 
\textbf{1)} We introduce TopoAlign, a novel method addressing the shortage of training corpora for Math LLMs by structurally aligning code with formal mathematical languages. 
\textbf{2)} We propose CAF, a training task that leverages the structurally aligned code dataset to emulate autoformalisation, thereby reducing the dependence on annotated pairs of informal and formal mathematical statements.
\textbf{3)} We release a large-scale semi-synthetic pre-training dataset of 300 million tokens, consisting of high-quality, structurally aligned code designed for autoformalisation tasks.
\textbf{4)} Through detailed ablation studies, we demonstrate that a balanced ratio of our aligned code data and formal mathematical statements yields optimal autoformalisation performance.

\section{Related Work}

Previous methods for autoformalisation draw inspiration from machine translation literature \citep{wang2018experimentsneuraltranslationinformal, dwivedi2022graphneuralnetworkslearnable}, i.e. \citet{szegedy2020promising} propose encoding NL and FL in a shared latent space and selecting translation candidates based on embedding similarity. Some approaches focus on rule-based methods, such as GFLean, which uses the Grammatical Framework for parsing and linearization \citep{pathak2024gfleanautoformalisationframeworklean}. However, these methods struggle to adapt to diverse inputs, as their rules require frequent updates. In contrast, LLMs provide more flexibility and consequently show strong autoformalisation performance \citep{jiang2022autoformalization, jiang2024language, weng2025autoformalization}.

Despite their success in narrow domains \citep{soroco2025pdecontrollerllmsautoformalizationreasoning,zhu2024fgeo}, these methods face a common challenge: the scarcity of paired NL-FL math datasets. 
Various approaches are aimed to extend the training datasets: ATLAS \citep{liu2025atlasautoformalizingtheoremslifting} employs a student-teacher framework to generate synthetic data, but its performance is bounded by the teacher model's mathematical knowledge. Additionally, ATLAS generates fully synthetic data from existing Lean code. Herald Statements \citep{gao2024herald} are synthetically generated and limit the novelty of the resulting data. \citet{jiang2024multilingual} demonstrate that high-quality multilingual data yields further performance gains in autoformalization, a finding echoed by \citet{chan2025leaning}. Several high-quality, human-annotated datasets have been curated, including ProofNet \citep{azerbayev2023proofnet}, \minif \citep{zheng_minif2f}, \putnam \citep{tsoukalas2024putnambenchevaluatingneuraltheoremprovers}, and the Mathlib library itself \citep{The_mathlib_Community_2020}. While invaluable, creating these datasets is resource-intensive, requires domain experts, and is consequently limited in scale.
To address this, we find code data as a scalable alternative to mathematical data.

This is motivated by \citet{chen2021evaluatinglargelanguagemodels}, who demonstrate that pretraining on code transfers to autoformalization, with Codex achieving notable few-shot performance on such tasks. As such, Math LLMs are initialised from LLMs trained on extensive code data and progressively fine-tuned on mathematical datasets. Llemma \citep{Zhang_2024}, Kimina \citep{wang2025kimina} and \dsmath \citep{shao2024deepseekmath} follow this paradigm. \citet{li2024autoformalize} argue that the autoformalization potential of Math LLMs remains underexplored when leveraging general-purpose code data during pretraining. To address this, we propose topologically decomposing and aligning code from repositories with formal math statements, unlocking a scalable new source of training data for LLMs.

Finally, given data scarcity, previous work explored techniques for efficient usage of available data. Related methods aim to extract the inherent relationships between NL and FL in the data, proposing alignment methods based on symbolic equivalence and semantic consistency \citep{li2024autoformalize}. However, aligning NL and FL remains challenging when their formats and structures differ, making it difficult to transfer the LLMs knowledge and reasoning capabilities between NL and FL.

\section{Methodology}

We posit that current Math LLM performance is constrained primarily by the scarcity of large-scale training corpora. Large code datasets have already proven valuable for initializing these models \citep{shao2024deepseekmath, wang2025kimina}, yet code data remains largely untapped during subsequent training on formal mathematics. Our TopoAlign framework and CAF address this gap, detailed as follows.

\subsection{Topological Decomposition of Code for Structural Alignment with Formal Math}

TopoAlign builds on the premise that code and formal mathematical statements share a compositional structure. Functions in code solve distinct subproblems and may rely on auxiliary functions, analogous to how formal statements resolve informal problem descriptions using lemmata and theorems. We therefore decompose code at the function level (see Figure~\ref{fig:parallel}) into three transferable components: (i) the docstring, corresponding to the informal problem statement, (ii) the main function body, serving as a proxy for the formal statement; and (iii) its dependency functions, corresponding to supporting lemmata or library theorems (i.e., from Mathlib in Lean~4). We select Lean as the representative mathematical formal language in this work, while other systems such as Isabelle \citep{isabelle} and Coq \citep{coq} are discussed in Appendix \ref{app:other math}. Disassembling code into components that mirror those in formal mathematics enables structural transfer. Importantly, TopoAlign targets structural alignment and does not assume semantic equivalence between programming-language type systems and Lean's dependent type system. The framework provides complementary pretraining data that \textbf{enhances, rather than replaces}, training on formal mathematical corpora.

\begin{figure}[h]
\centering
  \includegraphics[width=1.0\columnwidth]{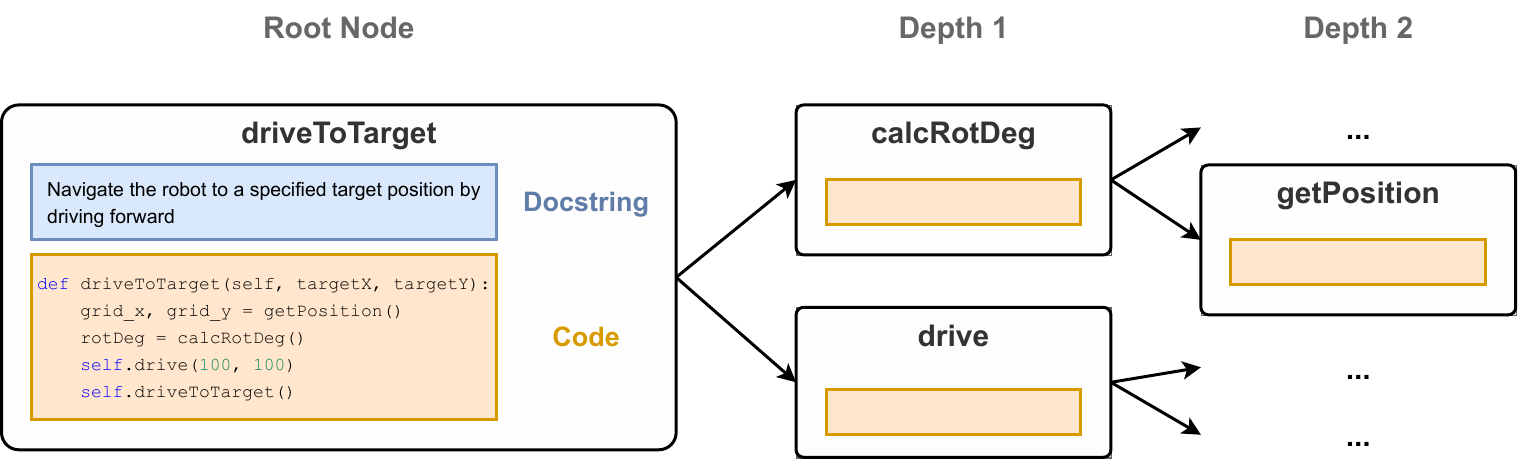}
  \caption{Function-level dependency tree showing the hierarchy of function calls, starting from the root node. Each child node represents a function called by its parent. The docstring for the root node is extracted to represent the description of the problem addressed.}
  \label{fig:sample}
  \vspace{-0.3cm}
\end{figure}

To extract functional dependencies, we employ a topological dependency parser that performs a breadth-first search to build function-level dependency graphs (see Algorithm~\ref{al:dependency} in Appendix~\ref{sec:appendix_algo}). This contrasts with file-level dependency extraction (i.e., DeepSeek \citep{guo2024deepseek}), which captures inter-file execution order but omits intra-file functional relationships. Our parser leverages abstract syntax trees to trace calls and parent definitions across files, and is designed to handle standalone functions, class or instance methods, recursive calls, and imports. The result is a tree-structured representation of code dependencies, as shown in Figure~\ref{fig:sample}. 

To obtain informal problem statements, we extract natural language descriptions from docstrings and README files. To obtain informal problem statements, we extract natural language descriptions from docstrings and README files. However, as docstrings describe a function's interface rather than its implementation, they do not always align with the needs of autoformalization. We therefore use \qwen \citep{qwen3} to generate concise summaries of each function's implementation logic, augmenting missing or low-quality documentation.
We perform 10-gram contamination analysis, following \citet{guo2024deepseek}, to ensure no information leakage between our dataset and the test set, and confirm no meaningful overlap.\footnote{Additional details are provided in Appendix \ref{app:prompts}.} We have also provided additional analyses of the generated dataset, including its length distribution and LLM-as-a-Judge scores, in Appendix \ref{sec:appendix_quality}.

\begin{figure}[h]
    \centering
    \includegraphics[width=0.9\linewidth]{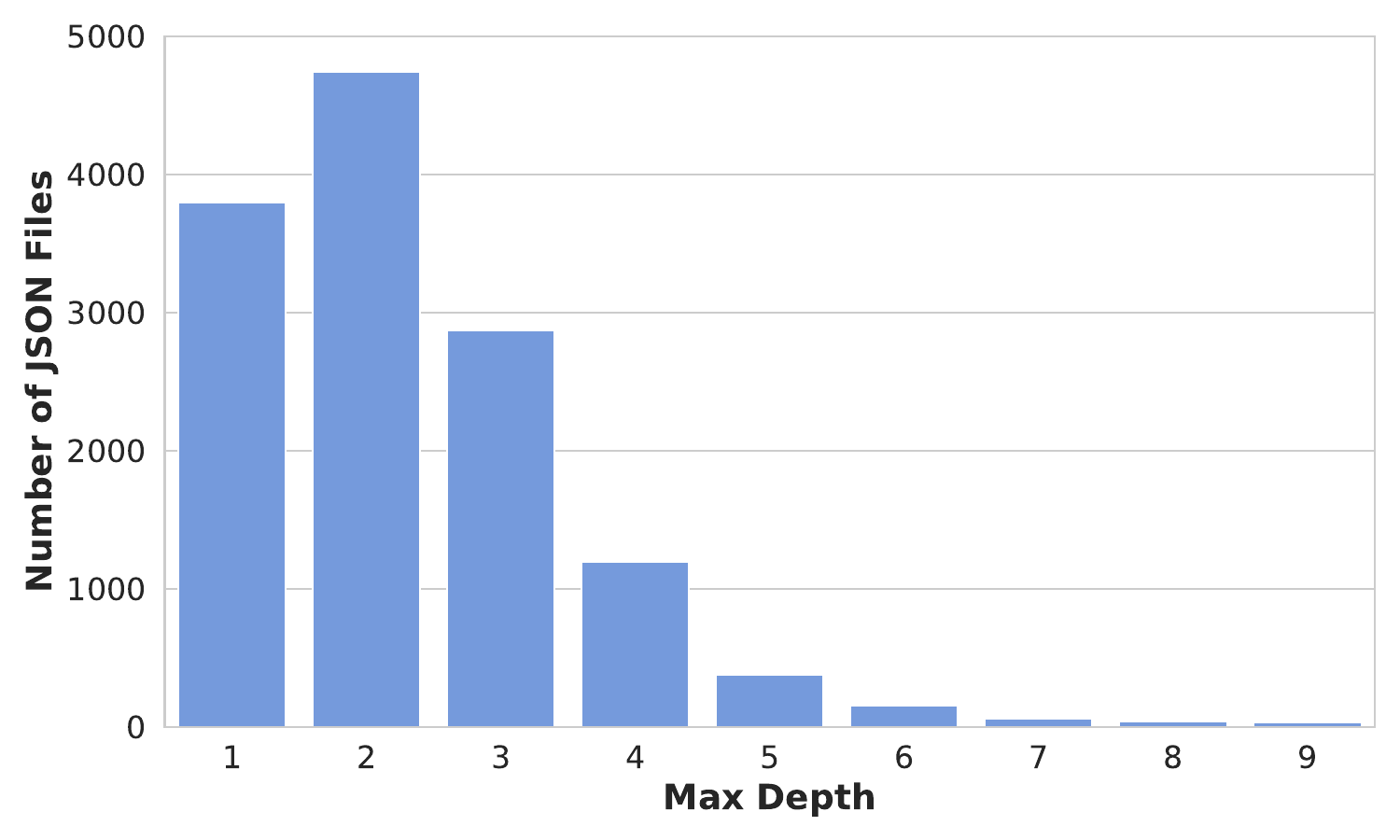}
    \caption{Distribution of maximum dependency tree depths across a random sample of 200 repositories.}
    \label{fig:data-depth}
\end{figure}

\begin{figure}[h]
    \centering
    \includegraphics[width=0.9\linewidth]{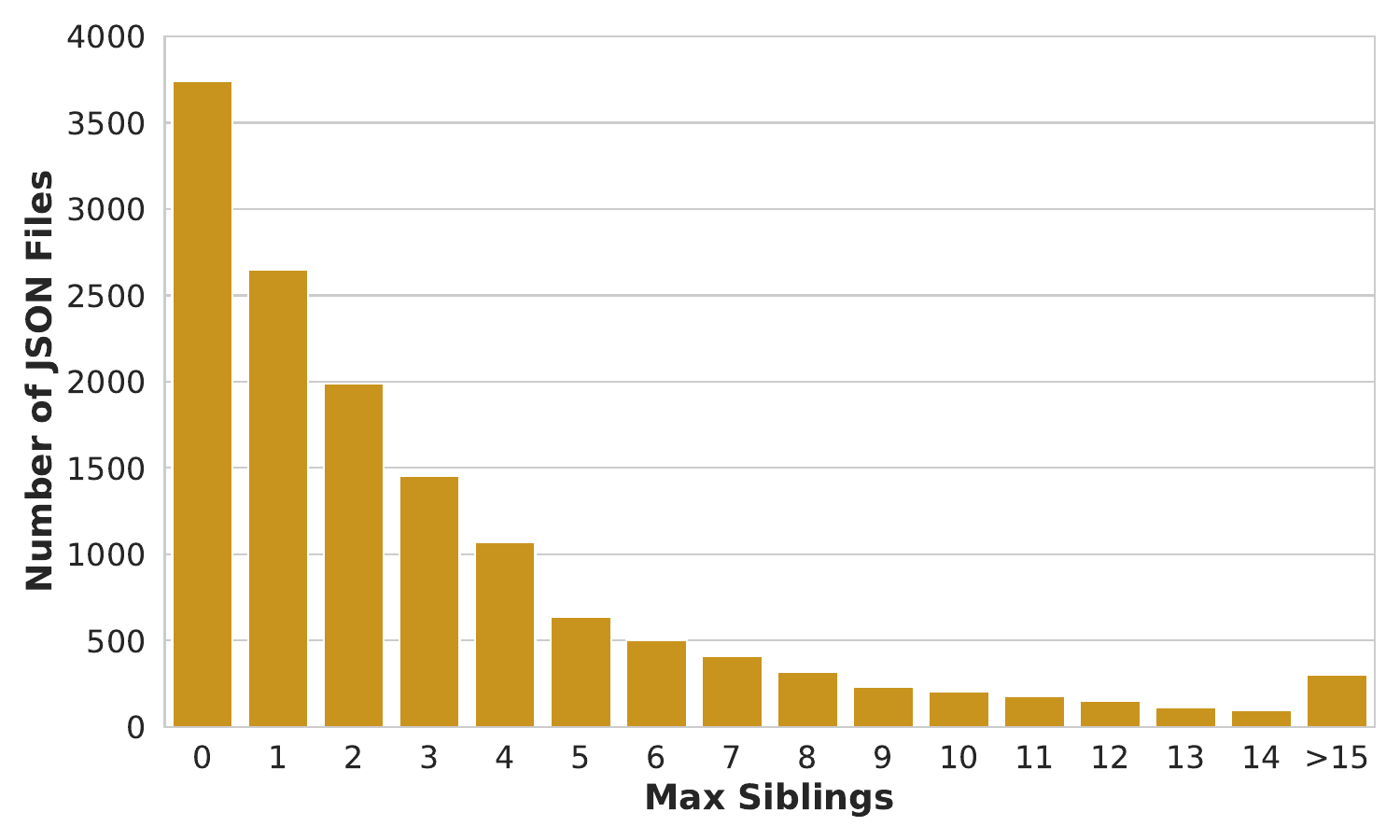}
    \caption{Distribution of the maximum number of siblings in dependency trees across a random sample of 200 repositories.}
    \label{fig:data-siblings}
    \vspace{-0.7cm}
\end{figure}

We analyse the structural properties of function-level dependency trees in the collected code to select repositories whose hierarchical patterns resemble those found in formal mathematics. For each repository we compute the maximum tree depth, reflecting overall complexity and the maximum number of sibling nodes at any depth, indicating the breadth of direct dependencies. We retain repositories with depth between 3 and 6 and maximum sibling counts between 3 and 10, thus excluding overly complex codebases as well as simple scripts. Figures~\ref{fig:data-depth} and~\ref{fig:data-siblings} show the distributions of maximum depth and maximum sibling count for a random sample of 200 repositories. After filtering, the corpus contains 156,684 functions comprising 324.5 million tokens.\footnote{Dataset available at: \url{https://huggingface.co/datasets/Formal-Math-Reasoning/TopoAlign_Python}.}

\begin{figure*}
    \centering
    \includegraphics[width=0.85\linewidth]{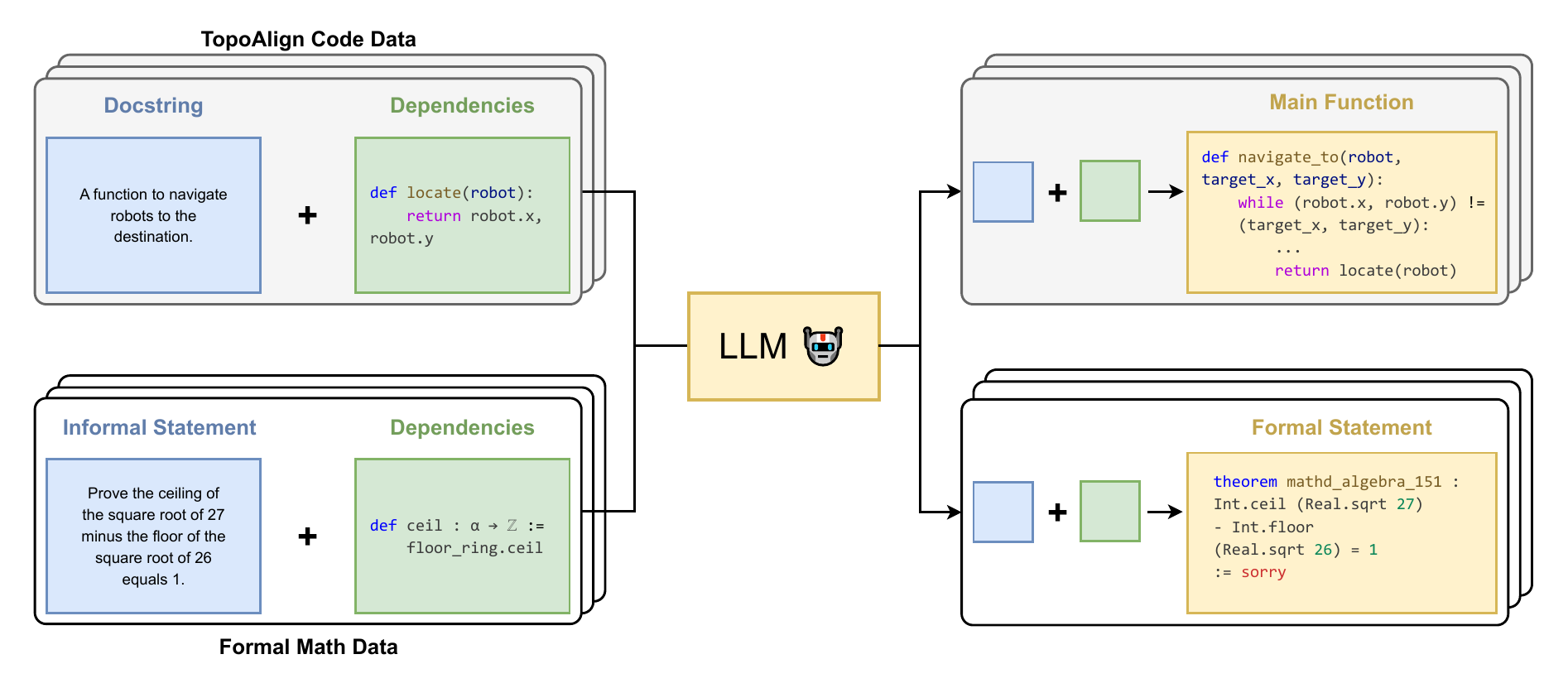}
    \caption{Overview of the training pipeline. Input consists of a problem description (for either code or math) and its dependencies. The training objective is to generate the corresponding solution: the root code block for code inputs, or the formal statement for math inputs.}
    \label{fig:pipeline}
    \vspace{-0.5cm}
\end{figure*}

\subsection{Code Autoformalisation (CAF)}

To leverage structurally aligned code for training Math LLMs on autoformalisation, we introduce CAF, which emulates the autoformalisation process on code data. Treating each aligned code instance as an analogue of a formalisation scenario allows the model to learn structural patterns while transferring general problem‑solving strategies acquired in programming. Importantly, our method focuses on transferring structural and problem-solving capabilities, rather than introducing new mathematical knowledge. While we recognize that this technique could be adapted to emulate other formal reasoning tasks like theorem proving, in this work, we focus specifically on addressing the prevalent data scarcity bottleneck in autoformalisation.

We train Math LLMs using a mixture of TopoAlign code data and formal mathematical data. This combined approach integrates mathematical knowledge and structure with problem-solving capabilities from code, while also mitigating catastrophic forgetting during fine-tuning \citep{chen2019catastrophic}. In our multi-task training approach, each training sample consists of an input $x$, a docstring for code or an informal statement for math, and a set of dependencies $d$, a set of dependency functions for code or supporting lemmata and theorems for math. The model is conditioned on $x$ and $d$ and trained to generate the target $y$: the main function for code data or the formal statement for math data, as illustrated in Figure~\ref{fig:pipeline}.

The proportion of code and math samples during training is controlled by parameter $\alpha$, where $\alpha$ determines the fraction of math samples and $1-\alpha$ the fraction of code samples. The overall objective is defined as $\mathcal{L} = \alpha\, \mathcal{L}_{\text{math}} + (1-\alpha)\, \mathcal{L}_{\text{CAF}}$, where $\mathcal{L}_{\text{math}}$ and $\mathcal{L}_{\text{CAF}}$ denote the losses for math and code tasks, respectively. For each task, the loss is computed using next-token prediction, formulated as the negative log-likelihood $\mathcal{L} = - \sum_{i=1}^{N} \log P_\theta \big( y_i \mid y_{<i}, x \big)$, where $x$ is the input sequence, $y = (y_1, y_2, \dots, y_N)$ is the target sequence, and $P_\theta$ is the model with parameters $\theta$.

\section{Experiments}

\rparagraph{Training Data}
Our code data is sourced from Python repositories in the Stack v2 dataset \citep{lozhkov2024starcoder}. For formal mathematical statements, we use the Herald Statements corpus \citep{gao2024herald}, a synthetic dataset built from Mathlib \citep{The_mathlib_Community_2020}. It contains Lean statements similar in style and structure to those found in math pretraining corpora, while avoiding overlap with downstream evaluation benchmarks such as \minif, \proofnet, and \putnam. Dependency functions for the formal statements are extracted using the jixia library.\footnote{\url{https://github.com/frenzymath/jixia}}

\rparagraph{Models}
We evaluate three base models: \dsmath (7B) \citep{shao2024deepseekmath}, \herald (7B) \citep{gao2024herald} and \qwen (4B) \citep{qwen3}. We evaluate three base models: \dsmath (7B) \citep{shao2024deepseekmath}, \herald (7B) \citep{gao2024herald} and \qwen (4B) \citep{qwen3} with their instruct versions. \herald is specialised for autoformalization, having been trained on the synthetic Herald Statements data. In contrast, \dsmath is trained on a broader range of mathematical data from DeepSeek-Coder \citep{guo2024deepseek}, optimised for general mathematical problem-solving rather than explicit autoformalization. We additionally evaluate \qwen, a state-of-the-art general-purpose reasoning model pre-trained on large-scale mathematical corpora to enhance its reasoning capabilities. This setup allows us to compare a dedicated autoformalization model against models with broader mathematical and code understanding.\footnote{We do not benchmark StepFun-Formalizer \citep{wu2026stepfun} or Kimina-Autoformalizer \citep{wang2025kimina} due to benchmark contamination, as \minif, \proofnet, and \putnam are included in their training data.}

We should note that TopoAlign should be understood as a continual pre-training methodology for math autoformalisation from code repositories, intended for use in the pre- and mid-training phases of LLMs providing more structural information and problem understanding information. As such, it is orthogonal to fine-tuning-based approaches. It follows that  to meaningfully compare TopoAlign against fine-tuned SOTA formalizers, it would require to retrain the latter with TopoAlign data under identical conditions, which is beyond the scope of this study. Instead, we compare against baselines using different data distributions as follows, which we argue provides a fair and interpretable evaluation of TopoAlign's contribution.

\rparagraph{Settings}
For each base model, we evaluate four distinct variations, namely \textbf{Baseline}, \textbf{Math}, \textbf{Code} and \textbf{TopoAlign}. The \textbf{Baseline} setting evaluates the pretrained models directly on downstream tasks without any additional fine-tuning. The \textbf{Math} setting involves fine-tuning models exclusively on formal mathematical data from the Herald Statements corpus, which is equivalent to applying the CAF objective with a mixing ratio of $\alpha=1.0$. This allows us to assess the impact of training on purely mathematical data. In the \textbf{Code} setting, models are trained only on unaligned code data extracted from the Stack-V2 corpus, providing a control to test the effect of training without structural alignment via TopoAlign. Finally, the \textbf{TopoAlign} setting trains models on a balanced combination of formal mathematics and structurally aligned code using CAF, with a mixing ratio of $\alpha=0.5$. This setting uses 4,000 randomly selected samples from GitHub repositories and 4,000 from the Herald Statements dataset. For consistency, the number of training samples is kept equal across all settings.\footnote{Each instance is a complete reasoning example and therefore cannot be meaningfully split to match token counts. Matching token counts by selecting shorter examples would instead bias the data toward simpler code or proofs.} Hyperparameter and prompt details are provided in Appendix~\ref{al:prompts}.

\rparagraph{Benchmarks}
We evaluate models on several Lean 4 autoformalisation benchmarks: \putnam \citep{tsoukalas2024putnambenchevaluatingneuraltheoremprovers}, the validation and test sets of \minif \citep{zheng_minif2f}, and \proofnet \citep{azerbayev2023proofnet}. Each benchmark consists of paired natural language and formal language statements. Among these, \putnam is the most challenging problems, \proofnet comprises textbook theorems, and \minif is considered the most in-distribution benchmark, as it overlaps with Mathlib.

\rparagraph{Evaluation Metrics}
We measure model performance using two primary metrics. First, we use Typecheck (TC) with the Lean 4 compiler (v4.21.0) to verify the syntactic correctness of generated statements \citep{poiroux2025improvingautoformalizationusingtype, limperg2025tactic, rabe2020mathematical, weng2025autoformalization}. For a more rigorous assessment of semantic fidelity, we employ bidirectional equivalence (BEq) \citep{liu2025rethinking}, which uses an LLM to generate proof tactics establishing logical equivalence between the model's output and the reference statement. This provides a stronger signal of faithful autoformalization than typechecking alone, and is consistent with the established LLM-as-judge evaluation paradigm \citep{gu2024survey}.

\section{Results and Discussion}
 
Table~\ref{tab:results} presents the Typecheck and BEq results under the pass@k metric \cite{chen2021codex} with $k=\{1,10\}$, where a problem is considered solved if at least one of the $k$ sampled solutions is correct, evaluated on the \minif-valid, \minif-test, \proofnet, and \putnam datasets. We additionally report unbiased pass@1 results in Appendix~\ref{sec:unbiased_pass}.

\input{tables/final_result}

Our proposed TopoAlign method outperforms all benchmarked methods across most of the datasets and model families on the BE@10 metric, with the exception of \herald on \proofnet. Table \ref{tab:avg_rank} shows that TopoAlign achieves the best overall model performance, measured by average rank across datasets, for \dsmath, \qwen, and \herald. For example, \dsmath BEq@1 on \minif-valid increases by 9.65\% to 14.47\%. Some variance in BEq@1 is expected due to stochastic sampling, which can account for occasional drops, such as the slightly lower score for \herald on \putnam. In the more robust BEq@10 evaluation, TopoAlign yields consistent improvements across \dsmath, \qwen, and \herald on \putnam, \minif-valid, and \minif-test. These results demonstrate that topological alignment effectively transfers structural knowledge from code to formal mathematics even, though it introduces no additional mathematical knowledge. These gains are observed across model architectures and datasets.

TopoAlign consistently surpasses code-only variants for both \herald and \dsmath on BEq and Typecheck, indicating that unaligned code data alone does not provide structural information beneficial for autoformalization. This is further evidenced by \herald, which already has more math knowledge than other base models, trained solely on code data. Its BEq scores and typecheck accuracy decreases significantly, with generated outputs frequently syntactically invalid despite occasional semantic plausibility (see Appendix~\ref{app:generated samples}). This suggests that raw code data not only fails to benefit autoformalization but may actively degrade an already capable model's formalization ability, highlighting the necessity of structural alignment and a proper training paradigm. In contrast, topological alignment and the CAF task enable effective transfer of both problem-solving and structural knowledge to formal mathematics. Notably, \qwen trained on raw code frequently generates degenerate outputs such as endless type definitions (e.g.\, Nat $\to$ Nat), which pass typechecking but are semantically meaningless, suggesting prior exposure to code typechecking during training.

These findings confirm that TopoAlign and CAF effectively leverage code data to enhance Math LLM training, addressing the scarcity of formal mathematical corpora by unlocking code repositories as a scalable source of structurally aligned pretraining data. Principally, TopoAlign unlocks code repositories as structurally aligned pretraining data while leaving the underlying code content unchanged, and CAF then facilitates the transfer of this knowledge from code to mathematics. 

\subsection{Qualitative Error Analysis}
To better understand model behavior, we conduct a qualitative error analysis on 200 failure cases, comprising 50 randomly selected samples per dataset generated by \qwen with TopoAlign. We categorize the errors into four distinct types as follows, with representative cases detailed in Appendix~\ref{app:error_analysis}.

\textbf{Type 1: Missing or Incorrect Conditions,} i.e. 
conditions are omitted or improperly specified, such as missing universal quantifiers or weakened constraints. \textbf{Type 2: Semantically Equivalent but Structurally Different,} i.e. 
generated statements are semantically correct but differ in structural representation, such as using set membership in place of disjunctive equality. \textbf{Type 3: Variable Type Mismatch,} i.e. 
variables are assigned incorrect or incompatible types, producing statements that are syntactically plausible but semantically invalid. \textbf{Type 4: Representation Inconsistency,} i.e. 
variables or expressions are inconsistently represented, leading to semantic mismatches, such as conflating absolute value notation with distance functions.

The distribution of error types across datasets is summarised in Table~\ref{tab:error_analysis}. The majority of errors fall under Type 1, accounting for approximately 70\% of all failures. The primary bottleneck is deemed as deep problem comprehension.
\input{tables/error_analysis}

This distribution also explains the uneven performance gains across datasets. For harder benchmarks such as ProofNet and Putnam, Type 1 errors overwhelmingly dominate, averaging over 90\% of failures, indicating that implicit natural language constraints are too complex for the model to parse reliably. Since TopoAlign provides structural alignment rather than additional mathematical knowledge, models with stronger inherent reasoning capabilities benefit more substantially, as they can actively leverage the aligned structure. Weaker models, by contrast, remain bottlenecked by their inability to comprehend the underlying math.

\subsection{Impact of Code–Math Data Ratio}
We further study the effect of varying the ratio of code to mathematical data in the CAF objective by adjusting the $\alpha$ parameter. We train additional \dsmath models with $\alpha \in {0.25, 0.75}$ and report the results in Table~\ref{tab:model_comparison}. A balanced ratio ($\alpha = 0.5$) consistently achieves the highest or near-highest performance for both TC@1 and BEq@1. Reducing the proportion of mathematical data ($\alpha = 0.25$) leads to the weakest performance, likely due to the token-level loss being dominated by the longer code samples, biasing the model toward code generation. Increasing the proportion of mathematical data ($\alpha = 0.75$) improves TC@1 scores, particularly on benchmarks closer to Mathlib, such as \minif, but does not consistently improve BEq@1. These findings indicate a balanced mix of code and mathematical data is crucial for optimal autoformalisation performance: math data enhances syntactic accuracy, while code enhances problem-solving capabilities.

\input{tables/ablation_alpha}

\subsection{Base model performance}
The \dsmath base model, which lacks pretraining on autoformalisation tasks, fails to generate meaningful outputs, often producing repeated symbols or malformed syntax that result in typecheck failures. In contrast, \herald, which is pretrained on a large mathematical corpus, provides strong BEq performance across multiple datasets. Notably, the introduction of CAF training further improves its results: for instance, BEq@1 on \minif-valid increases from 23.07\% to 25.08\%, and on \minif-test from 22.70\% to 25.64\%. These improvements confirm that TopoAlign provides significant benefits even for models that already possess strong autoformalisation capabilities.

\section{Conclusion}
We show that widely available code repositories provide a valuable and largely untapped source of training data for mathematical LLMs. We introduce TopoAlign, a structural alignment framework that maps code to formal mathematics and yields a 324.5 million token dataset that mirrors the structure of formal mathematical statements. Training on this dataset substantially improves performance across four autoformalization benchmarks on both Typecheck and BEq, demonstrating that structural and problem-solving knowledge can be transferred from code to formal mathematics through structural alignment. Our ablations further show that a balanced mix of code and formal mathematical data is important for achieving the best performance. Beyond autoformalization, TopoAlign may also provide a useful foundation for related tasks such as theorem proving.

\section*{Limitations}

Due to computational and resource constraints, our main experiments focus on a subset of approximately 4,000 code samples and 4,000 math samples. We would like to emphasize 324.5M tokens are currently only considering code repositories in Python (around 28\% in Stack-V2). This corpus can be further extended using other programming languages in Stack-V2.

Moreover, while TopoAlign is designed to augment and structurally enrich existing training data rather than replace high-quality mathematical corpora, its performance on research-level benchmarks such as FATE-H \cite{jiang2025fate} remains limited, with all models yielding less than BEq of 0.05. This reflects an inherent constraint: structural alignment cannot substitute for the advanced mathematical knowledge that only high-quality formal datasets can provide, and such datasets remain scarce at the research level. Additionally, TopoAlign cannot replace high-quality data. For example, our additional experiments with Goedel-Formalizer, which was trained on Mathlib, show that continued pre-training with additional mathematical and TopoAlign data does not improve BEq performance (Appendix~\ref{app:goedel}). This further highlights the importance of the quality and complementarity of the data used for continued pre-training. These findings are consistent with the intended scope of TopoAlign, which is to improve the structural quality and utility of available training data rather than expand the model's underlying mathematical knowledge.

\section*{Ethic consideration}
This research focuses exclusively on computational methods for mathematical reasoning and the development of machine learning frameworks to improve autoformalisation. It does not involve human subjects, personal data, or any intervention in real-world populations. All experiments are conducted on publicly available datasets or synthetic data derived from code repositories, and no sensitive or private information is used.

\bibliography{custom}

\appendix

\section{Additional results}
\label{sec:appendix}
\subsection{Unbiased results}
\label{sec:unbiased_pass}

We report the unbiased results in Table~\ref{tab:unbiased_results} as supplementary evidence supporting our main findings. These results, presented alongside prior literature for direct comparison, show a trend where TopoAlign appears to outperform other training frameworks in terms of BEq score and Typecheck. Due to resource constraints, we reported pass@1, but they still provide supporting evidence for our hypothesis.
\input{tables/unbiased_final}
\subsection{Rank}
\label{sec:rank}
We also provide a rank-based comparison in Table~\ref{tab:avg_rank}, statistically summarized via the standard pass@k metric, to offer a concise overview across different training settings. This is more interpretable than the raw numbers reported in Table~\ref{tab:unbiased_results} and Table~\ref{tab:results}. From the average-rank perspective, TopoAlign consistently achieves an average rank between 1 and 2, outperforming all other settings, which further corroborates the effectiveness of TopoAlign.
\input{tables/rank}
\section{Software and Data}
To ensure full reproducibility, all data sources, tools, and code are publicly available. All three basemodels we select from huggingface checkpoint (\url{https://huggingface.co/deepseek-ai/deepseek-math-7b-instruct}, \url{https://huggingface.co/Qwen/Qwen3-8B}, \url{https://huggingface.co/FrenzyMath/Herald_translator}).The downstream evaluation benchmarks were sourced directly from their official repositories: \minif (\url{https://github.com/openai/miniF2F}), \putnam (\url{https://github.com/trishullab/PutnamBench}), and Herald (\url{https://huggingface.co/datasets/FrenzyMath/Herald_statements}). Ground truth dependencies were processed using Jixia (\url{https://github.com/frenzymath/jixia}), as detailed in Section 4. Our complete codebase and the generated dataset have been released at \url{https://huggingface.co/datasets/Formal-Math-Reasoning/TopoAlign_Python}.

\section{Training Hyperparameters}
\label{al:prompts}
The hyperparameters used for training the LLMs are listed in Tables \ref{tab:hyperparameters_1} and \ref{tab:hyperparameters_2}. We adhere to the default settings provided in the original papers. In our setup, we fine-tune the base models using supervised fine-tuning. The implementation is based on the Hugging Face Transformers library (version 4.52).

\input{tables/hyperparameters}

\section{Generated Sample from Code-Only Trained Model}
\label{app:generated samples}
These samples are generated by \herald trained exclusively on code data. The outputs demonstrate that while the model captures the underlying logical structure correctly, it lacks proficiency in Lean 4 syntax. However, this suggests the presence of semantic similarity between programming code and formal mathematical language.

\begin{mycodebox}
from math import sqrt, ceil, floor

def evaluate_sqrt_expressions_1():
    assert(int(ceil(sqrt(27))) - int(floor(sqrt(26))) == 1)

def is_increasing_function(a, b):
    '''
    The function checks if the function f(x) = 4bx + (a+1)x^2 is increasing for x >= 0.
    '''
    return 4*b <= 4*b**2 + (a+1)**2

def is_3_(girls: list[int]) -> int:
    """
    >>> is_3_([1,2,3,4,5,6,7])
    3
    """
    return 3
\end{mycodebox}

\section{Error Analysis Examples}
\label{app:error_analysis}

\paragraph{Type 1: Missing or Incorrect Conditions.}
The generated output losses the universal quantification over all positive $k$ and all integer multiples $m$. The hypothesis \texttt{h\textsubscript{1} : 0 < n} is also entirely omitted.

\begin{myleanbox}
\underline{Reference:}\\
\texttt{h\textsubscript{0} : $\forall$ k : $\mathbb{N}$, 0 < k $\to$ $\forall$ m : $\mathbb{Z}$, x $\neq$ m * $\pi$ / (2\^{}k)}\\
\texttt{h\textsubscript{1} : 0 < n}\\[6pt]
\underline{Generated:}\\
\texttt{hx : x $\neq$ $\pi$ / (2\^{}(n+1))}
\end{myleanbox}
\paragraph{Type 2: Semantically Equivalent but Structurally Different.}
The generated output uses set membership notation $\in \{\cdots\}$ in place of explicit disjunctive equality. While the two expressions are semantically equivalent, they differ structurally, which may cause mismatches during formal verification or proof search.

\begin{myleanbox}
\underline{Reference:}\\
\texttt{(x, y) = (1, 1) $\vee$ (x, y) = (16, 2) $\vee$ (x, y) = (27, 3)}\\[6pt]
\underline{Generated:}\\
\texttt{(x, y) $\in$ \{ (1, 1), (16, 2), (27, 3) \}}
\end{myleanbox}

\paragraph{Type 3: Variable Type Mismatch.}
The generated output assigns an integer type $\mathbb{Z}$ to a variable that should be typed as a natural number $\mathbb{N}$. Although syntactically plausible, this type mismatch renders the statement semantically invalid within the formal type system.

\begin{myleanbox}
\underline{Reference:}\\
\texttt{n : $\mathbb{N}$}\\[6pt]
\underline{Generated:}\\
\texttt{n : $\mathbb{Z}$}
\end{myleanbox}

\paragraph{Type 4: Representation Inconsistency.}
The reference uses \texttt{abs} for absolute value and \texttt{Real.sqrt} with explicit coordinate arithmetic for Euclidean distance. The generated output instead applies the \texttt{dist} function inconsistently: once as a unary distance from zero (conflating absolute value), and once as a binary point distance, mixing two distinct representations within the same statement.

\begin{myleanbox}
\underline{Reference:}\\
\texttt{abs y = 6}\\
\texttt{Real.sqrt ((x - 8)\^{}2 + (y - 3)\^{}2)}\\[6pt]
\underline{Generated:}\\
\texttt{dist x 0 = 6}\\
\texttt{dist (x, y) (8, 3)}
\end{myleanbox}
\newpage
\section{Training Instructions and Inference Prompts}
\label{app:prompts}

To generate docstrings describing the main functions in the code repositories, we employ Qwen3-14B model (\url{https://huggingface.co/Qwen/Qwen3-14B}) using the following prompt format:

\begin{mypromptbox}
Provide a concise description of the problem solved in the code snippet below. Format the response as a docstring.

\{code\}
\end{mypromptbox}

We manually verified 50 randomly sampled docstrings from our dataset, and all of them exhibit accurate and complete descriptions of the corresponding code. Selected samples are provided below:

\begin{mycodebox}
def sync(self, title: str, keywords: Union[str, list[str]]) -> dict:
    
    title = validate_title(title)
    keywords = validate_keywords(keywords)
    filter_item = self.filter(title)
    remote_keywords = filter_item["keywords"]
    [...]
    params = OrderedDict(
        {
            "title": title,
            "context[]": filter_item["context"],
            "filter_action": filter_item["filter_action"],
        }
    )
    params.update(self._keyword_params(add_key + delete_key))
    response = self._call_api(
        "put", f"/api/v2/filters/{filter_item['id']}", params=params
    )
    response["added"] = add_keywords
    response["deleted"] = delete_keywords
    return response

\end{mycodebox}

\begin{docstringbox}
This function synchronizes local and remote keywords for a filter by comparing provided keywords with existing ones. It identifies keywords to add and remove, logs changes, constructs API parameters, and updates the filter via a PUT request. The response includes added/deleted keywords to confirm synchronization.
\end{docstringbox}
\begin{mycodebox}
def attention_mlp_stack(inputs, out_channel, num_heads, depth, key_dim, attn_ratio, mlp_ratio, strides, stack_drop=0, activation="hard_swish", name=""):

    nn = inputs
    embed_dim = nn.shape[-1]
    stack_drop_s, stack_drop_e = stack_drop if isinstance(stack_drop, (list, tuple)) else [stack_drop, stack_drop]
    for ii in range(depth):
        block_name = name + "block{}_".format(ii + 1)
        drop_rate = stack_drop_s + (stack_drop_e - stack_drop_s) * ii / depth
        nn = res_mhsa_with_multi_head_position(nn, embed_dim, num_heads, key_dim, attn_ratio, drop_rate, activation=activation, name=block_name)
    
    if mlp_ratio > 0:
    [...]
    if mlp_ratio > 0:
        nn = res_mlp_block(nn, mlp_ratio, drop_rate, activation=activation, name=block_name + "mlp_")
    
    return keras.layers.Activation("linear", name=name + "output")(nn)

\end{mycodebox}

\begin{docstringbox}
This function constructs a neural network stack with multi-head attention and MLP blocks. It applies a series of attention mechanisms (res\_mhsa\_with\_multi\_head\_position) and optional MLPs (res\_mlp\_block) across `depth' layers, adjusting dropout rates progressively. The stack supports configurable heads, key dimensions, ratios, and activation functions, returning a linear output layer.
\end{docstringbox}

These examples demonstrate that generated docstrings faithfully capture the code: the first enhances a minimal existing docstring with concise descriptions of keyword synchronization logic, while the second provides a detailed description of the multi-layer attention–MLP architecture for a function that originally lacked a docstring.

For math-related tasks, we use the following prompt format:
\begin{mypromptbox}

Use the following pre-defined Lean 4 dependencies: \\
\{dependencies\}\\
\\
Based on the context and the problem description, generate a single, syntactically correct Lean 4 formal statement that accurately captures the problem's meaning.\\
\\
Problem Description:\\
\{problem description\}

\end{mypromptbox}
For code-related tasks, we employ a parallel prompt structure:
\begin{mypromptbox}

Use the following pre-defined functions:\\
\{pre-defined functions\}\\
\\
Based on the context and the problem description, generate a syntactically correct function implementation that accurately captures the problem's meaning.
\\
Problem Description:\\
\{problem description\}

\end{mypromptbox}

\section{Quality of generated dataset}
\label{sec:appendix_quality}

We provide an extended quality analysis of TopoAlign along three dimensions:  length, repetition, and semantic quality. As shown in Table~\ref{tab:topoalign_length},  the generated docstrings contain an average of 356.74 tokens, 52.97 words, and 2.54 sentences, with relatively small variance in the median statistics. The maximum values are higher due to a small number of substantially longer docstrings. As shown in Table~\ref{tab:topoalign_repetition}, the dataset exhibits negligible repetition, with 250,734 unique lines and a unique ratio of 99.99\%; only 0.02\% of docstrings have an exact-match duplicate.

\begin{table}[h]
\centering
\caption{Length statistics of the generated docstrings in TopoAlign.}
\label{tab:topoalign_length}
\resizebox{\columnwidth}{!}{
\begin{tabular}{lrrrr}
\toprule
\textbf{Dimension} & \textbf{Mean} & \textbf{Std.} & \textbf{Median} & \textbf{Max} \\
\midrule
Token number    & 356.74 & 87.14 & 353 & 4,391 \\
Word number     & 52.97  & 12.48 & 52  & 870 \\
Sentence number & 2.54   & 1.04  & 2   & 19 \\
\bottomrule
\end{tabular}}
\end{table}

\begin{table}[h]
\centering
\caption{Repetition statistics of the generated docstrings in TopoAlign.}
\label{tab:topoalign_repetition}
\resizebox{0.9\columnwidth}{!}{

\begin{tabular}{lr}
\toprule
\textbf{Metric} & \textbf{Value} \\
\midrule
Unique lines & 250,734 \\
Unique ratio & 99.99\% \\
Exact-match repeated lines & 0.02\% \\
\bottomrule
\end{tabular}}
\end{table}

To further assess the semantic quality of the generated docstrings, we randomly sample 10,000 examples and evaluate them using three independent LLM judges, DeepSeek-V4-Pro, Qwen3.7-Max, and Gemini-3.5-Flash. As reported in Table~\ref{tab:topoalign_llm_judge}, all three judges consistently assign high scores for correctness, completeness, and hallucination-free generation. The strongest scores are achieved by Gemini-3.5-Flash, with average scores of 4.91, 4.81, and 4.94 out of 5 for correctness, completeness, and hallucination-free generation, respectively, while the other two judges show 
the same overall trend.

\begin{table}[h]
\centering
\caption{LLM-as-a-Judge evaluation of 10,000 randomly selected generated docstrings from TopoAlign. Scores range from 1 to 5, with higher scores indicating better quality.}
\label{tab:topoalign_llm_judge}
\resizebox{\columnwidth}{!}{

\begin{tabular}{lccc}
\toprule
\textbf{Model} & \textbf{Correctness} & \textbf{Completeness} & \textbf{Hallucination-free} \\
\midrule
Gemini-3.5-Flash & $4.91 \pm 0.35$ & $4.81 \pm 0.40$ & $4.94 \pm 0.28$ \\
Qwen3.7-Max     & $4.73 \pm 0.60$ & $4.39 \pm 0.63$ & $4.83 \pm 0.52$ \\
DeepSeek-V4-Pro  & $4.41 \pm 0.96$ & $3.68 \pm 0.87$ & $4.45 \pm 1.04$ \\
\bottomrule
\end{tabular}}
\end{table}

The prompts are shown below for the LLM-as-a-judge.

\begin{mypromptbox}
\textbf{[SYSTEM]} You are a rigorous code-summary evaluator. You will be given a Python function (with its surrounding dependency code) and a natural-language docstring that claims to describe what the function does.

Rate the docstring on three axes on an INTEGER scale of 1--5:

\textbf{1. Correctness} --- Do the statements in the docstring accurately reflect the function's actual behavior? Penalize wrong claims about control flow, return values, side effects, or algorithm.
(1 = mostly wrong, 5 = fully accurate)

\textbf{2. Completeness} --- Does the docstring cover the function's key behavior: inputs, outputs, main control flow / algorithmic steps, side effects, and error handling? Penalize missing pieces.
(1 = large gaps, 5 = covers all salient behavior)

\textbf{3. Hallucination} --- Does the docstring introduce facts NOT supported by the code (invented behavior, non-existent parameters, wrong algorithm)? HIGHER score = LESS hallucination.
(1 = many invented claims, 5 = no unsupported claims)

Return ONLY a JSON object, no prose, no markdown:
\texttt{\{"correctness": <int 1-5>, "completeness": <int 1-5>, "hallucination": <int 1-5>, "reason": "<one sentence, <=40 words>"\}}

\medskip
\textbf{[USER]}

\texttt{\#\#\# Dependency code (context, may be empty)}

\begin{verbatim}
\{DEPENDENCY\}
\{MAIN\}
\{DOCSTRING\}
\end{verbatim}
\end{mypromptbox}

\section{Dependency Tree Extraction Algorithm}
\label{sec:appendix_algo}
The dependency tree extraction process is formally described in Algorithm \ref{al:dependency}. The algorithm iterates through all files within a given directory, systematically identifying user-defined functions, including both class methods and standalone functions. We then analyse inter-file and inter-class dependency relationships. Through static analysis of function call patterns, we construct a directed dependency graph where each function call establishes a parent-child relationship. The calling function serves as the parent node, while the called function becomes the child node.
\begin{algorithm*}[h!]
\caption{Function Call Dependency Analysis}
\label{al:dependency}
\begin{algorithmic}[1]
\State \textbf{Input:} Python project directory
\State \textbf{Output:} Dependency graph of function calls

\State \textbf{Data Structures:}
\State $\mathit{function\_definitions} \gets \{\}$ \Comment{Record defined functions}
\State $\mathit{imports} \gets \{\}$ 
$\mathit{object\_types} \gets \{\}$ \Comment{Imported names $\to$ full paths; Object names $\to$ class names; Save these two to track their parent function definitions or the file where they're defined}

\State \Call{AnalyzeFile}{$f$} for all Python files $f$ in directory (recursively)

\Procedure{AnalyzeFile}{file\_path}
    \State Parse AST from $file\_path$
    \State Initialize $\mathit{function\_calls} \gets \{\}$, $\mathit{object\_types} \gets \{\}$
    
    \ForAll{nodes $n$ in AST}
        \If{$n$ is \textbf{ClassDef}}
            \State Track current class context\Comment{Class defined function}
        \ElsIf{$n$ is \textbf{Import/ImportFrom}}
            \State Record $\mathit{imports}[alias] \gets$ full module path\Comment{Function from other files}
        \ElsIf{$n$ is \textbf{FunctionDef}}
            \State Register function with $\mathit{current\_class}.name$ if applicable
            \State Initialize $\mathit{function\_calls}[name] \gets []$\Comment{Save its dependency}
        \ElsIf{$n$ is \textbf{Assign} with constructor call}
            \State Map $\mathit{object\_types}[var] \gets$ class name\Comment{Record the codes that initialise this function}
        \ElsIf{$n$ is \textbf{Call}}\Comment{Record the codes that call this function}
            \State Resolve full function name (direct call($obj()$), method call($obj.method()$), or imports $\mathit{imports}[name]$
           
            \State Append to $\mathit{function\_calls}[\mathit{current\_func}]$\Comment{The list to record the dependency information}
        \EndIf
    \EndFor
\EndProcedure

\Procedure{BuildDependencyGraph}{}
    \State Filter to keep only project-internal calls
    \State Detect recursive calls ($f \to f$)
    \Comment{Self-recursive call}
    
    \State \textbf{Topological Sort:}\Comment{BFS search}

    \State Construct nested call tree from sorted order
    \State Insert recursion markers at tree root if needed
\EndProcedure

\end{algorithmic}
\end{algorithm*}

\section{Extending TopoAlign to other  Formal Languages}
\label{app:other math}

Formal statements in other proof assistants (e.g., Coq and Isabelle) share structural similarities with Lean. Below we show Isabelle and Coq versions of the informal statement: \emph{Prove the ceiling of the square root of 27 minus the floor of the square root of 26 equals 1.}

\begin{myleanbox}
\underline{Isabelle:}

theorem ceil\_sqrt27\_minus\_floor\_sqrt26:
  ``ceiling (sqrt 27) - floor (sqrt 26) = (1::int)"
\end{myleanbox}

\begin{myleanbox}
\underline{Coq:}

Theorem ceil\_sqrt27\_minus\_floor\_sqrt26 :
  (up (sqrt 27) - Z.floor (sqrt 26))\%Z = 1\%Z.
\end{myleanbox}

While these formal languages show similar structural characteristics to the decomposed code data in TopoAlign, supporting multiple formal languages would require either training dedicated models for each target language or constructing a mixed pretraining corpus that incorporates multiple formal languages. Although this direction offers promising practical applications, it lies beyond the scope of the present work.

\section{Additional Ablation: Goedel-Formalizer}
\label{app:goedel}

We additionally evaluate Goedel-Formalizer-V2-8B \cite{lin2025goedelproverfrontiermodelopensource} under the same experimental settings as our main experiments, and report the results separately in Table~\ref{tab:biased_results_goedel}. Across all continued pre-training settings, BEq performance decreases relative to the baseline, although Typecheck performance occasionally improves. This suggests that additional mathematical or code data does not necessarily improve downstream autoformalization quality and can instead reduce semantic correctness as measured by BEq. Among the continued pre-training settings, TopoAlign achieves the highest BEq@10 performance on \minif-test, \proofnet, and \putnam, although it remains below the baseline. These results also highlight the importance of carefully selecting the mathematical data used for continued pre-training: the Math setting yields the lowest BEq@10 performance on \minif-valid, \minif-test, and \putnam, suggesting that additional formal mathematical data can reduce formalization performance when it is not sufficiently complementary to the model's existing training distribution.

\begin{table}[htbp]
\caption{Autoformalization performance (\%) of Goedel-Formalizer-V2-8B \cite{lin2025goedelproverfrontiermodelopensource} on \minif, \proofnet, and \putnam, reporting pass@1 and pass@10 for Typecheck (TC) and BEq.}
\label{tab:biased_results_goedel}
\resizebox{0.48\textwidth}{!}{%
\begin{tabular}{c l llll}
\toprule
\textbf{Dataset} & \textbf{Setting} & \textbf{TC@1} & \textbf{BEq@1} & \textbf{TC@10} & \textbf{BEq@10} \\
\midrule
\multirow{4}{*}{\textbf{\minif-valid}} 
& \cellcolor{gray!15}Baseline  & \cellcolor{gray!15}80.70 & \cellcolor{gray!15}23.87 & \cellcolor{gray!15}80.70 & \cellcolor{gray!15}24.32 \\
& Code   & 65.79 & \textbf{20.74} & 65.79  & \textbf{20.74} \\
& Math   & \textbf{83.77} & 17.98 & \textbf{83.77} & 18.42 \\
& TopoAlign & 80.70 & 19.74 & 81.14 & 20.61 \\
\midrule
\multirow{4}{*}{\textbf{\minif-test}} 
& \cellcolor{gray!15}Baseline  & \cellcolor{gray!15}84.38 & \cellcolor{gray!15}22.97 & \cellcolor{gray!15}84.38  & \cellcolor{gray!15}23.87 \\
& Code   & 70.09 & 18.72 & 70.54 & 19.25 \\
& Math   & \textbf{83.93} & 17.41 & \textbf{83.93} & 17.86 \\
& TopoAlign & 81.25 & \textbf{19.20} & 81.70 & \textbf{20.54} \\
\midrule
\multirow{4}{*}{\textbf{\proofnet}} 
& \cellcolor{gray!15}Baseline  & \cellcolor{gray!15}51.07 & \cellcolor{gray!15}7.40 & \cellcolor{gray!15}51.07 & \cellcolor{gray!15}7.67 \\
& Code   & 50.00 & 3.48 & 50.00 & 3.49 \\
& Math   & \textbf{53.21} & 5.88 & \textbf{53.21} & 5.88 \\
& TopoAlign & 52.14 & \textbf{6.43} & 52.41 & \textbf{6.43} \\
\midrule
\multirow{4}{*}{\textbf{\putnam}} 
& \cellcolor{gray!15}Baseline  & \cellcolor{gray!15}53.14 & \cellcolor{gray!15}3.41 & \cellcolor{gray!15}53.14  & \cellcolor{gray!15}3.75 \\
& Code   & 41.82 & \textbf{1.36} & 45.28 & 1.36 \\
& Math   & \textbf{60.38} & 0.63 & \textbf{60.38} & 0.94 \\
& TopoAlign & 58.81 & 1.26 & 59.43 & \textbf{1.57} \\
\bottomrule
\end{tabular}}
\end{table}

\end{document}

%% file: math_commands.tex
\usepackage{amsmath,amsfonts,bm}

\def\eqref#1{equation~\ref{#1}}

\def\1{\bm{1}}

\DeclareMathAlphabet{\mathsfit}{\encodingdefault}{\sfdefault}{m}{sl}
\SetMathAlphabet{\mathsfit}{bold}{\encodingdefault}{\sfdefault}{bx}{n}



%% file: variables.tex
\usepackage{xspace}
\usepackage{mdwlist}
\usepackage{tabularx}
\usepackage{booktabs}
\usepackage{colortbl}

\newcommand{\rparagraph}[1]{\vspace{1.4mm}\noindent\textbf{#1.}}

\newcommand{\dsmath}{\textsc{DeepSeek-Math}\xspace}
\newcommand{\herald}{\textsc{Herald}\xspace}
\newcommand{\qwen}{\textsc{Qwen-3}\xspace}
\newcommand{\minif}{MiniF2F\xspace}
\newcommand{\proofnet}{ProofNet\xspace}
\newcommand{\putnam}{Putnam\xspace}

\newcolumntype{Y}{>{\centering\arraybackslash}X}

%% file: tables/final_result.tex
\begin{table*}[t!]
\begin{center}
\caption{Auto-formalisation performance in percent for \minif, \proofnet, and \putnam datasets under pass@1 and pass@10 metrics for Typecheck (TC) and BEq. Baseline setting refers to the pretrained model, Math is trained on additional formal math data and code is trained on additional code data that is not structurally aligned. {TopoAlign} mixes math and structurally aligned code data.}
{
\def\arraystretch{1.1}
\fontsize{9.0pt}{9.2pt}\selectfont
\resizebox{0.9\textwidth}{!}{
\begin{tabularx}{\textwidth}{cc l XX XX}
\toprule
\textbf{Dataset} & \textbf{Base Model} & \textbf{Setting} & \textbf{TC@1} & \textbf{BEq@1} & \textbf{TC@10} & \textbf{BEq@10} \\
\midrule

\multirow{12}{*}{\textbf{\minif-valid}} 
& \multirow{4}{*}{\dsmath} 
    & \cellcolor{gray!15}Baseline &  \cellcolor{gray!15} \phantom{0}0.00 &  \cellcolor{gray!15} \phantom{0}0.00 &  \cellcolor{gray!15} \phantom{0}0.00 &  \cellcolor{gray!15} \phantom{0}0.00 \\
    & & Math        & 45.18 & \phantom{0}9.65 & 79.82     & 19.74     \\
    & & Code        & 42.98 & 10.53 & \textbf{88.16}     & 23.68     \\
    & & TopoAlign        & \textbf{51.75} & \textbf{14.47} & 87.28     & \textbf{26.32}    \\
\cmidrule{2-7}
& \multirow{4}{*}{\qwen} 
    & \cellcolor{gray!15}Baseline 
& \cellcolor{gray!15}  \phantom{0}2.63 
& \cellcolor{gray!15}  \phantom{0}0.00 
& \cellcolor{gray!15} 21.93 
& \cellcolor{gray!15}  \phantom{0}0.00 \\
& & Math      
& 81.58 
& \textbf{13.16} 
& \textbf{100.00} 
& 82.02 \\
& & Code      
& \textbf{97.37} 
& \phantom{0}0.00 
& \textbf{100.00} 
& \phantom{0}0.00 \\
& & TopoAlign 
& 77.19 
& \textbf{13.16} 
& \textbf{100.00} 
& \textbf{85.53} \\

\cmidrule{2-7}
& \multirow{4}{*}{\herald}
    & \cellcolor{gray!15}Baseline   & \cellcolor{gray!15} 73.25 & \cellcolor{gray!15} 25.44 & \cellcolor{gray!15} 92.98     & \cellcolor{gray!15} 38.16     \\
    & & Math    & 75.44 & 24.12 & 93.86     & \textbf{41.23}     \\
    & & Code    & 75.00 & 20.61 & 90.79     & 32.02     \\
    & & TopoAlign & \textbf{76.75} &  \textbf{26.75} & \textbf{94.74} & \textbf{41.23}\\

\midrule

\multirow{12}{*}{\textbf{\minif-test}} 
& \multirow{4}{*}{\dsmath} 
    & \cellcolor{gray!15}Baseline &  \cellcolor{gray!15} \phantom{0}0.00 &  \cellcolor{gray!15} \phantom{0}0.00 &  \cellcolor{gray!15} \phantom{0}0.00 &  \cellcolor{gray!15} \phantom{0}0.00 \\
    & & Math         & 43.11 & \phantom{0}9.33 & 77.33     & 21.78     \\
    & & Code       & 52.89 & \phantom{0}7.56 &  \textbf{91.52}    & 26.79     \\
    & & TopoAlign        & \textbf{54.67} & \textbf{16.89} & 88.89     & \textbf{29.78}     \\
\cmidrule{2-7}
& \multirow{4}{*}{\qwen} 

& \cellcolor{gray!15}Baseline 
& \cellcolor{gray!15}  \phantom{0}6.22 
& \cellcolor{gray!15}  \phantom{0}0.00 
& \cellcolor{gray!15} 37.33 
& \cellcolor{gray!15}  \phantom{0}0.00 \\
& & Math      
& 84.44 
& 10.22 
& \textbf{100.00} 
& 74.22 \\
& & Code      
& \textbf{97.78} 
& \phantom{0}0.00 
& \textbf{100.00} 
& \phantom{0}0.00 \\
& & TopoAlign 
& 88.89 
& \textbf{23.11} 
& \textbf{100.00} 
& \textbf{84.00} \\
\cmidrule{2-7}

& \multirow{4}{*}{\herald}
    & \cellcolor{gray!15}Baseline   & \cellcolor{gray!15} 78.22 & \cellcolor{gray!15} 24.44 & \cellcolor{gray!15} \textbf{95.54}     & \cellcolor{gray!15} 41.96     \\
    & & Math    & \textbf{80.89} & 24.44 & \textbf{95.54}     & 40.36     \\
    & & Code    & \textbf{80.89} & 20.89 & 90.63 & 34.38 \\
    & & TopoAlign & 79.56 & \textbf{27.56} & 94.20 & \textbf{42.86}\\

\midrule

\multirow{12}{*}{\textbf{\proofnet}} 
& \multirow{4}{*}{\dsmath} 
    & \cellcolor{gray!15}Baseline & \cellcolor{gray!15} \phantom{0}0.00 &  \cellcolor{gray!15} \phantom{0}0.00 &  \cellcolor{gray!15} \phantom{0}0.00 &  \cellcolor{gray!15} \phantom{0}0.00 \\
    & & Math         & 21.12 & \phantom{0}5.35 & 43.42     & 12.57     \\
    & & Code         & 22.73 & \phantom{0}2.67 & 49.47     & \phantom{0}7.49     \\
    & & TopoAlign         & \textbf{32.89} & \textbf{\phantom{0}9.09} & \textbf{56.95}    & \textbf{14.97}     \\
\cmidrule{2-7}
& \multirow{4}{*}{\qwen} 

& \cellcolor{gray!15}Baseline 
& \cellcolor{gray!15}  \phantom{0}6.42 
& \cellcolor{gray!15}  \phantom{0}0.16 
& \cellcolor{gray!15} 31.02 
& \cellcolor{gray!15}  \phantom{0}0.80 \\
& & Math      
& 28.34 
& \phantom{0}\textbf{5.88} 
& 88.77 
& 37.70 \\
& & Code      
& \textbf{56.15} 
& \phantom{0}0.00 
& \textbf{99.47} 
& \phantom{0}0.00 \\
& & TopoAlign 
& 31.02 
& \phantom{0}4.55 
& 95.72 
& \textbf{41.71} \\
\cmidrule{2-7}

& \multirow{4}{*}{\herald}
    & \cellcolor{gray!15}Baseline & \cellcolor{gray!15} 46.52 & \cellcolor{gray!15} 10.16 & \cellcolor{gray!15} 74.87     & \cellcolor{gray!15} \textbf{20.32}     \\
    & & Math    & \textbf{46.79} & \textbf{10.43} & 75.40    & 19.77    \\
    & & Code    & 38.24 & \phantom{0}4.55 & 63.37    & \phantom{0}9.36    \\
    & & TopoAlign & 43.85 & \phantom{0}9.63 & \textbf{75.67}  & 16.84\\

\midrule

\multirow{12}{*}{\textbf{\putnam}} 
& \multirow{4}{*}{\dsmath} 
    & \cellcolor{gray!15}Baseline & \cellcolor{gray!15} \phantom{0}0.00 &  \cellcolor{gray!15} \phantom{0}0.00 &  \cellcolor{gray!15} \phantom{0}0.00 &  \cellcolor{gray!15} \phantom{0}0.00 \\
    & & Math         & 10.69 &  \phantom{0}0.00    & 27.04    &  \phantom{0}0.00     \\
    & & Code         & 12.58 &  \phantom{0}0.00   & 37.42    & \phantom{0}0.00    \\
    & & TopoAlign        & \textbf{17.30} &  \phantom{0}0.00    & \textbf{42.14}     &  \phantom{0}0.00     \\
\cmidrule{2-7}
& \multirow{4}{*}{\qwen} 

& \cellcolor{gray!15}Baseline 
& \cellcolor{gray!15}  \phantom{0}1.89 
& \cellcolor{gray!15}  \phantom{0}0.00 
& \cellcolor{gray!15} 16.98 
& \cellcolor{gray!15}  \phantom{0}0.00 \\
& & Math      
& 25.79 
& \phantom{0}0.63 
& 98.11 
& \phantom{0}5.35 \\
& & Code      
& \textbf{70.44} 
& \phantom{0}0.00 
& \textbf{100.00} 
& \phantom{0}0.00 \\
& & TopoAlign 
& 19.50 
& \phantom{0}\textbf{1.26} 
& 97.48 
& \phantom{0}\textbf{9.12} \\
\cmidrule{2-7}

& \multirow{4}{*}{\herald}
    & \cellcolor{gray!15}Baseline   & \cellcolor{gray!15} 37.42 & \cellcolor{gray!15} \phantom{0}2.20 & \cellcolor{gray!15} 73.58     & \cellcolor{gray!15} \textbf{\phantom{0}4.72}     \\
    & & Math    & \textbf{43.71} & \textbf{\phantom{0}2.52} & 70.44 & \phantom{0}4.09     \\
    & & Code    & 35.85 & \phantom{0}0.00 & 69.18 & \phantom{0}0.00     \\
    & & TopoAlign & 36.16 & \phantom{0}1.57 & \textbf{76.73} & \textbf{\phantom{0}4.72}\\

\bottomrule
\end{tabularx}
}}
\label{tab:results}
\vspace{-12pt}
\end{center}
\end{table*}

%% file: tables/error_analysis.tex
\begin{table}[h]
\centering
\caption{Distribution of error types across datasets (50 samples per dataset) generated by \qwen.}
\label{tab:error_analysis}
\resizebox{\columnwidth}{!}{
\begin{tabular}{lcccccc}
\toprule
\textbf{Dataset} & \textbf{Type 1} & \textbf{Type 2} & \textbf{Type 3} & \textbf{Type 4} & \textbf{Other} \\
\midrule
\minif-valid & 28 & 14 & 5 & 2 & 1 \\
\minif-test  & 23 & 15 & 5 & 4 & 3 \\
\proofnet      & 43 &  2 & 1 & 1 & 3 \\
\putnam        & 47 &  1 & 2 & 0 & 0 \\
\bottomrule
\end{tabular}}
\end{table}

%% file: tables/ablation_alpha.tex
\begin{table}[tbp]
\begin{center}
\caption{Ablation study on training data composition for \dsmath. The table compares Typecheck (TC) and BEq scores for pass@1 to identify the optimal ratio of formal math data to our aligned code data.}
{
\def\arraystretch{1.1}
\fontsize{8.0pt}{8.2pt}\selectfont
\begin{tabular}{c l c c}
\toprule
\textbf{Dataset} & \textbf{Data ratio} & \textbf{TC@1} & \textbf{BEq@1} \\
\midrule
\multirow{3}{*}{\minif-valid} 
    & $\alpha = 0.25$ & 44.74 & 11.84 \\
    & $\alpha = 0.50$ & 51.75 & \textbf{14.47} \\
    & $\alpha = 0.75$ & \textbf{52.63} & 10.96 \\
\midrule
\multirow{3}{*}{\minif-test} 
    & $\alpha = 0.25$ & 42.67 & \phantom{0}7.56 \\
    & $\alpha = 0.50$ & 54.67 & \textbf{16.89} \\
    & $\alpha = 0.75$ & \textbf{56.44} & 14.67 \\
\midrule
\multirow{3}{*}{\putnam} 
    & $\alpha = 0.25$ & 12.58 & \phantom{0}0.00 \\
    & $\alpha = 0.50$ & \textbf{17.30} & \phantom{0}0.00 \\
    & $\alpha = 0.75$ & 12.89 & \phantom{0}0.00 \\
\midrule
\multirow{3}{*}{\proofnet} 
    & $\alpha = 0.25$ & 24.60 & \phantom{0}6.15 \\
    & $\alpha = 0.50$ & \textbf{32.89} & \textbf{\phantom{0}9.09} \\
    & $\alpha = 0.75$ & 26.20 & \phantom{0}8.29 \\

\bottomrule

\end{tabular}
\vspace{-0.5cm}

}
\label{tab:model_comparison}
\end{center}
\end{table}

%% file: tables/unbiased_final.tex
\begin{table}[t!]
\begin{center}
\caption{Auto-formalisation performance in percent for \minif, \proofnet, and \putnam datasets under \textbf{unbiased} pass@1 metrics for Typecheck (TC) and BEq. Baseline setting refers to the pretrained model, Math is trained on additional formal math data and code is trained on additional code data that is not structurally aligned. {Topoalign} mixes math and structurally aligned code data.}

\resizebox{0.48\textwidth}{!}{%
\begin{tabular}{cc l ll}
\toprule
\textbf{Dataset} & \textbf{Base Model} & \textbf{Setting} & \textbf{TC@1} & \textbf{BEq@1} \\
\midrule

\multirow{12}{*}{\textbf{\minif-valid}} 
& \multirow{4}{*}{\dsmath} 
& \cellcolor{gray!15}Baseline & \cellcolor{gray!15} 2.19 & \cellcolor{gray!15} 0.00 \\
& & Math      & 38.16 & 10.92 \\
& & Code      & 36.44 & 12.41 \\
& & TopoAlign & \textbf{44.92} & \textbf{15.90} \\

& \multirow{4}{*}{\qwen} 
& \cellcolor{gray!15}Baseline & \cellcolor{gray!15} 2.19 & \cellcolor{gray!15} 0.00 \\
& & Math      & 78.46 & \textbf{13.95} \\
& & Code      & \textbf{97.37} & 0.00 \\
& & TopoAlign & 77.68 & \textbf{15.31} \\

& \multirow{4}{*}{\herald}
& \cellcolor{gray!15}Baseline & \cellcolor{gray!15} 78.34 & \cellcolor{gray!15} 23.07 \\
& & Math      & 75.74 & 24.69 \\
& & Code      & 78.59 & 21.49 \\
& & TopoAlign & \textbf{78.38} & \textbf{25.08} \\

\midrule

\multirow{12}{*}{\textbf{\minif-test}} 
& \multirow{4}{*}{\dsmath} 
& \cellcolor{gray!15}Baseline & \cellcolor{gray!15} 0.00 & \cellcolor{gray!15} 0.00 \\
& & Math      & 46.40 & 7.91 \\
& & Code      & 48.89 & 7.47 \\
& & TopoAlign & \textbf{55.83} & \textbf{15.61} \\

& \multirow{4}{*}{\qwen} 
& \cellcolor{gray!15}Baseline & \cellcolor{gray!15} 4.13 & \cellcolor{gray!15} 0.00 \\
& & Math      & 84.00 & 13.33 \\
& & Code      & \textbf{98.67} & 0.00 \\
& & TopoAlign & 79.60 & \textbf{17.33} \\

& \multirow{4}{*}{\herald}
& \cellcolor{gray!15}Baseline & \cellcolor{gray!15} 77.78 & \cellcolor{gray!15} 22.70 \\
& & Math      & \textbf{81.02} & 21.64 \\
& & Code      & 80.84 & 21.29 \\
& & TopoAlign & 80.45 & \textbf{25.64} \\

\midrule

\multirow{12}{*}{\textbf{\proofnet}} 
& \multirow{4}{*}{\dsmath} 
& \cellcolor{gray!15}Baseline & \cellcolor{gray!15} 0.00 & \cellcolor{gray!15} 0.00 \\
& & Math      & 21.65 & 4.33 \\
& & Code      & 20.46 & 1.79 \\
& & TopoAlign & \textbf{33.72} & \textbf{8.32} \\

& \multirow{4}{*}{\qwen} 
& \cellcolor{gray!15}Baseline & \cellcolor{gray!15} 4.84 & \cellcolor{gray!15} 0.80 \\
& & Math      & 27.54 & 4.60 \\
& & Code      & \textbf{54.55} & 0.00 \\
& & TopoAlign & 30.03 & \textbf{4.63} \\

& \multirow{4}{*}{\herald}
& \cellcolor{gray!15}Baseline & \cellcolor{gray!15} 46.92 & \cellcolor{gray!15} 10.13 \\
& & Math      & \textbf{47.00} & 9.12 \\
& & Code      & 27.77 & 2.55 \\
& & TopoAlign & 46.63 & \textbf{9.84} \\

\midrule

\multirow{12}{*}{\textbf{\putnam}} 
& \multirow{4}{*}{\dsmath} 
& \cellcolor{gray!15}Baseline & \cellcolor{gray!15} 0.00 & \cellcolor{gray!15} 0.00 \\
& & Math      & 8.14 & 0.00 \\
& & Code      & 10.82 & 0.00 \\
& & TopoAlign & \textbf{16.26} & 0.00 \\

& \multirow{4}{*}{\qwen} 
& \cellcolor{gray!15}Baseline & \cellcolor{gray!15} 1.73 & \cellcolor{gray!15} 0.00 \\
& & Math      & 29.84 & 0.57 \\
& & Code      & \textbf{69.18} & 0.00 \\
& & TopoAlign & 22.96 & \textbf{0.91} \\

& \multirow{4}{*}{\herald}
& \cellcolor{gray!15}Baseline & \cellcolor{gray!15} 41.98 & \cellcolor{gray!15} 2.76 \\
& & Math      & \textbf{43.87} & \textbf{2.87} \\
& & Code      & 34.84 & 0.00 \\
& & TopoAlign & 36.03 & 1.70 \\

\bottomrule
\end{tabular}%
}
\label{tab:unbiased_results}
\vspace{-12pt}
\end{center}
\end{table}

%% file: tables/rank.tex
\begin{table}[h]
\centering
\caption{Average rank across different settings.}
\label{tab:avg_rank}
\resizebox{0.9\columnwidth}{!}{
\begin{tabular}{l l c}
\toprule
\textbf{Model} & \textbf{Setting} & \textbf{Avg.Rank $\downarrow$} \\
\midrule

\multirow{4}{*}{\dsmath}
& Baseline  & 3.63 \\
& Math      & 2.50 \\
& Code      & 2.00 \\
& TopoAlign & \textbf{1.13} \\

\midrule

\multirow{4}{*}{\qwen}
& Baseline  & 3.50 \\
& Math      & 1.94 \\
& Code      & 2.13 \\
& TopoAlign & \textbf{1.63} \\

\midrule

\multirow{4}{*}{\herald}
& Baseline  & 2.25 \\
& Math      & 1.81 \\
& Code      & 3.75 \\
& TopoAlign & \textbf{1.69} \\

\bottomrule
\end{tabular}}
\end{table}

%% file: tables/hyperparameters.tex
\begin{table}[b]
\resizebox{0.9\columnwidth}{!}{

\begin{tabular}{lcc}
\toprule
\textbf{Hyperparameters} & \textbf{\dsmath} & \textbf{\herald} \\
\midrule
learning rate & 2e-4 & 1e-6\\
weight decay & 0.1 & 0.1 \\
epochs & 1 & 1 \\
batch size & 32 & 32 \\
\bottomrule
\end{tabular}
}
\caption{Training hyperparameter settings.}
\label{tab:hyperparameters_1}
\end{table}

\begin{table}[h!]
\centering
\resizebox{0.9\columnwidth}{!}{
\begin{tabular}{lcc}
\toprule
\textbf{Hyperparameters} & \textbf{\dsmath} & \textbf{\herald} \\
\midrule
Temperature & 0.70 & 0.70 \\
Top-$p$ & 0.95 & 0.95 \\
Max tokens & 1024 & 1024 \\
\bottomrule
\end{tabular}
}
\caption{Inference hyperparameter settings used for autoformalisation in the pass@10 setting.}
\label{tab:hyperparameters_2}
\end{table}